%% file: main.tex
\documentclass[lettersize,journal]{IEEEtran}
\usepackage{comment}
\usepackage{amsmath, amssymb}   
\usepackage{graphicx}           
\usepackage{color}              
\usepackage{pifont}             
\usepackage{booktabs}           
\usepackage{multirow}           
\usepackage{hyperref}           

\usepackage{times}  
\usepackage{helvet}  
\usepackage{courier}  
\usepackage{caption} 

\usepackage{algorithm}
\usepackage{algorithmic}
\usepackage{booktabs}
\usepackage{mathtools}
\usepackage{xcolor}
\DeclareMathOperator*{\argmin}{arg\,min}

\newcommand{\xmark}{\ding{55}}  
\newcommand{\cmark}{\ding{51}}  
\newcommand{\ours}{\texttt{OR-POSE}}

\renewcommand{\today}{\the\day\space\ifcase\month\or
  January\or February\or March\or April\or May\or June\or July\or
  August\or September\or October\or November\or December\fi\space\the\year}

\begin{document}

\title{Unsupervised Domain Adaptation for Occlusion Resilient Human Pose Estimation}

\author{Arindam Dutta$^{*}$,
Sarosij Bose$^{*}$,
Saketh Bachu,
Calvin-Khang Ta, 
Konstantinos Karydis,
Amit K. Roy-Chowdhury~\IEEEmembership{Fellow,~IEEE}
\thanks{$^{*}$ Equal contribution}
\thanks{Arindam Dutta, Sarosij Bose, Saketh Bachu, Konstantinos Karydis and Amit K. Roy-Chowdhury are with the Department of Electrical and Computer Engineering, University of California Riverside, USA. Calvin-Khang Ta is with the Department of Computer Science and Engineering, University of California Riverside, USA. 
}
\thanks{Corresponding authors: Arindam Dutta and Amit K. Roy-Chowdhury (email: adutt020@ucr.edu and amitrc@ece.ucr.edu).}
\thanks{Manuscript submitted \today.}}

\markboth{Journal of \LaTeX\ Class Files,~Vol.~14, No.~8, August~2021}%
{Shell \MakeLowercase{\textit{et al.}}: A Sample Article Using IEEEtran.cls for IEEE Journals}


\maketitle

\begin{abstract}
Occlusions are a significant challenge to human pose estimation algorithms, often resulting in inaccurate and anatomically implausible poses. Although current occlusion-robust human pose estimation algorithms exhibit impressive performance on existing datasets, their success is largely attributed to supervised training and the availability of additional information, such as multiple views or temporal continuity. Furthermore, these algorithms typically suffer from performance degradation under distribution shifts. While existing domain adaptive human pose estimation algorithms address this bottleneck, they tend to perform suboptimally when the target domain images are occluded, a common occurrence in real-life scenarios. To address these challenges, we propose \ours~: Unsupervised Domain Adaptation for \textbf{O}cclusion \textbf{R}esilient Human \textbf{POSE} Estimation. 
\ours~is an innovative unsupervised domain adaptation algorithm which effectively mitigates domain shifts and overcomes occlusion challenges by employing the mean teacher framework for iterative pseudo-label refinement. Additionally, \ours~reinforces realistic pose prediction by leveraging a learned human pose prior which incorporates the anatomical constraints of humans in the adaptation process. Lastly, \ours~avoids overfitting to inaccurate pseudo labels generated from heavily occluded images by employing a novel visibility-based curriculum learning approach. This enables the model to gradually transition from training samples with relatively less occlusion to more challenging, heavily occluded samples. Extensive experiments show that \ours~outperforms existing analogous state-of-the-art algorithms by $\sim$ \textbf{7}\% on challenging occluded human pose estimation datasets. 
\end{abstract}


\input{sections/1_intro}
\input{sections/2_related}
\input{sections/3_method}
\input{sections/4_experiments}
\input{sections/5_conclusion}

\bibliographystyle{IEEEtran}
\bibliography{main}

\end{document}

%% file: sections/1_intro.tex
\section{Introduction}

Human pose estimation is a key problem in computer vision which aims to precisely detect anatomical keypoints on the human body. 
Precise pose estimates are particularly crucial for real-world applications such as action recognition~\cite{zhang2024pgvt}, person re-identification~\cite{zhu2024seas} and gait recognition~\cite{teepe2021gaitgraph}. The remarkable success of state-of-the-art human pose estimation algorithms~\cite{cheng2020higherhrnet, xu2022zoomnas, yang2023effective} on challenging datasets~\cite{h36m_pami, lin2014microsoft} can be largely accredited to their access to extensive labeled training data. However, acquiring large amounts of real-world annotated data can be costly, time-consuming, and may raise privacy concerns~\cite{xiang2022being}. A trivial solution to these problems is to train models on large amounts of synthetic data and then apply these models to real-world datasets. However, models trained on synthetic data often suffer from performance degradation when evaluated on real-world data due to inherent domain differences between the two distributions, as demonstrated in~\cite{jiang2021regressive, kim2022unified}. Additionally, the presence of occlusions, a common phenomenon in real-life scenarios, continues to challenge the effectiveness of current pose estimation algorithms for practical usage~\cite{zhou2020occlusion}. 

\begin{figure}
    \centering
    \includegraphics[width=1\linewidth]{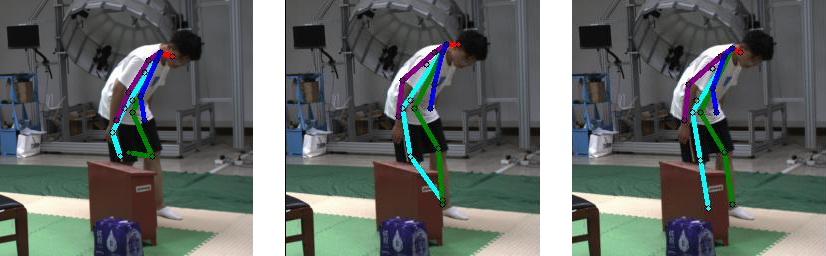}
    \caption{
    \textbf{Need for unsupervised domain adaptation for occlusion resilient human pose estimation.} \textit{Left:} Predictions of the model trained exclusively on labeled source data (SURREAL) and evaluated on an image from 3DOH50K dataset (referred to as {\it Source only} predictions). \textit{Middle:} Predictions from state-of-the-art domain adaptive human pose estimation algorithm UniFrame~\cite{kim2022unified}. \textit{Right:} Predictions from our proposed occlusion resilient algorithm (\ours). While UniFrame~\cite{kim2022unified} provides improved pose estimates as compared against {\it Source only} predictions through unsupervised domain adaptation to the unlabeled target domain, it still fails to deliver optimal pose estimates under occlusions. In contrast, our proposed algorithm (\ours) achieves optimal human pose estimates under unsupervised domain adaptive settings even in the presence of occlusions.} 
    \label{fig:teaser}
\end{figure}

Utilizing the capability of unsupervised domain adaptation (UDA) algorithms to transfer knowledge from a labeled source dataset to an unlabeled target dataset, several recent studies~\cite{jiang2021regressive,kim2022unified,peng2023source,raychaudhuri2023prior} addressed issues related to sub-optimal generalization of pose estimators under domain shifts, while also decreasing dependency on ground-truth annotations. These algorithms have shown significant success in achieving robust human pose estimation under unsupervised domain adaptive settings. However, these algorithms encounter observable challenges when dealing with occlusions in unlabeled target domain images, as demonstrated qualitatively in Figure~\ref{fig:teaser}. 

In addressing occlusion-based performance degradation, algorithms like those proposed in~\cite{zhou2020occlusion,cheng20203d,liu2022explicit} rely heavily on supervised learning techniques. For instance, \cite{zhou2020occlusion} requires paired occluded and unoccluded images to train a Siamese network for handling occlusions. \cite{cheng20203d} uses information from multiple frames to incorporate spatio-temporal continuity for managing self-occlusions in humans. Although these algorithms achieve notable performance improvements, they rely on large annotated training datasets and remain vulnerable to performance degradation when faced with distribution shifts~\cite{jiang2021regressive,kim2022unified}. Given these bottlenecks, we focus on addressing the specific problem of \emph{unsupervised domain adaptation for occlusion resilient human pose estimation}. 

\begin{figure}
    \centering
    \includegraphics[width=1\linewidth]{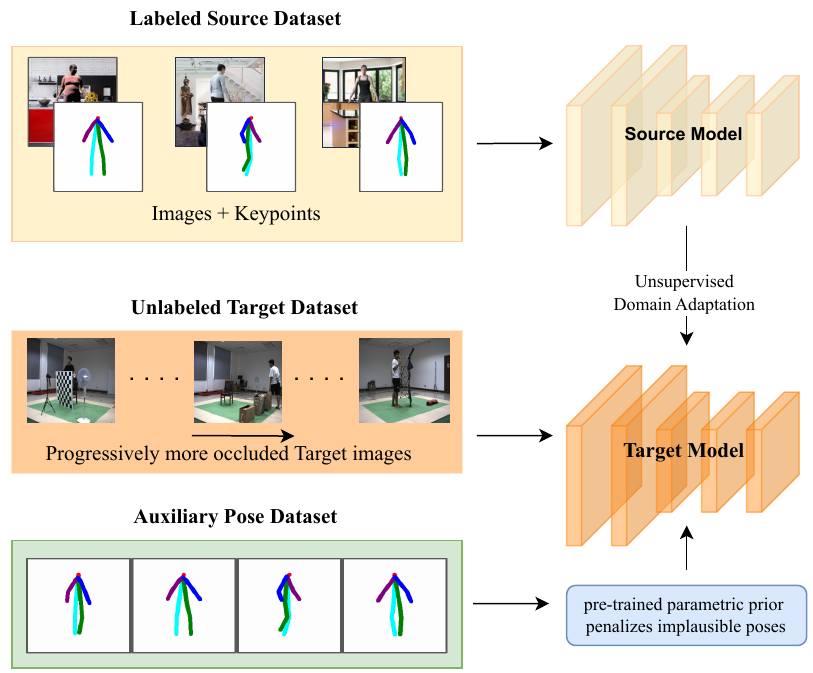}
    \caption{\textbf{Problem Setup.} We propose \ours, an unsupervised algorithm for progressively adapting a model to occlusions. \ours~leverages pseudo labels from the mean-teacher framework to provide guidance to the model while utilizing a pose prior to generate physically plausible poses for humans. Further, to prevent the early-overfitting to noisy pseudo labels and accounting for the uneven levels of occlusion present in the target domain, we suggest a curriculum learning based strategy to make the model learn from visible samples to harder (more occluded) samples.}
    \label{fig:overview}
\end{figure}

Our goal is to adapt a human pose estimation model trained on labelled source data to an unlabeled dataset of target images where human subjects are partially obscured by occlusions from inanimate objects. To address this challenge, we propose \ours: Unsupervised Domain Adaptation for \textbf{O}cclusion \textbf{R}esilient Human \textbf{POSE} Estimation, a novel self-training framework for addressing the aforementioned problem. \ours~generates improved pseudo labels from the teacher model for training of the student model by leveraging the popular Mean-Teacher framework~\cite{tarvainen2017mean} to enforce a consistency between the predictions of the student model and the teacher model.
By updating the teacher model's weights as an exponential moving average of the weights of the student model, the teacher model avoids overfitting to incorrect pseudo labels from the unlabeled target domain. Also, \ours~applies  occlusion augmentations~\cite{sarandi2020robust} on the labelled source data, allowing the student model and hence, the teacher model, to provide better pseudo labels by learning to generalize to occluded images. 

We observe that the above improvements serve to enhance the overall performance of \ours. However, due to the absence of human anatomical context, the predicted pose estimates remain suboptimal. 
Inspired by works such as~\cite{tiwari22posendf}, we develop a human pose prior that captures plausible human anatomy as a zero-level set. 
This prior regularizes the self-training algorithm by penalizing anatomically implausible predictions. Additionally, inspired by~\cite{karim2023c}, we introduce a novel visibility-based curriculum learning strategy. Initially, the model trains on more visible samples with fewer occlusions, then progressively shifts to less visible samples with higher occlusions. As occlusions are random in both nature and location,
this approach allows the network to learn from more reliable pseudo labels in the early stages of adaptation. Thus, this prevents error accumulation that typically results from learning from incorrect pseudo labels. Furthermore, \ours~works out-of-the-box for both occluded and non-occluded images, without requiring prior knowledge of the unlabelled target domain image. An overview of the proposed problem setup is provided in Figure~\ref{fig:overview}. \\


\noindent \textbf{Contributions:} We make the following key contributions in this work: \begin{itemize}
    \item We tackle the problem of unsupervised domain adaptation for occlusion-robust human pose estimation, addressing the current bottleneck of sub-optimal performance in existing UDA for human pose estimation algorithms when dealing with occlusions. 

    \item We introduce \ours, a novel self-training framework that leverages the mean-teacher approach and incorporates occlusion augmentations in the labelled source data, allowing the model to progressively refine its predictions on the unlabelled target domain.


    \item Comprehensive experiments on several challenging benchmarks show that \ours~significantly outperforms  existing algorithms (by $\sim$~7\%) on occluded, unlabeled target datasets, while maintaining performance on par with existing methods for unoccluded datasets. 

\end{itemize}

%% file: sections/2_related.tex
\section{Related Works}

\noindent \textbf{Human Pose Estimation:} 
Human pose estimation focuses on localizing anatomical landmarks on the human body across images and videos. Advancements in deep learning and the availability of large-scale datasets have significantly accelerated progress in 2D human pose estimation \cite{Pose_estimation_survey}. Existing algorithms, which generally consist of top-down~\cite{fang2017rmpe,he2017mask,xiao2018simple,sun2019deep} and bottom-up methods~\cite{cao2019openpose,cheng2020higherhrnet,jin2020differentiable,geng2021bottom}, typically rely on fully-supervised settings and often struggle to generalize to unseen domains.


When the human subject is occluded, pose estimation algorithms frequently struggle to accurately identify keypoints. To enable occluded keypoint detection, \cite{zhou2020occlusion} leveraged siamese networks and feature matching, which augmented unoccluded images to generate corresponding occluded images. \cite{cheng20203d} transferred knowledge from unoccluded to occluded frames using spatio-temporal continuity, leading to enhanced pose estimation results. Recently, \cite{liu2022explicit} proposed a multi-stage architecture designed to detect keypoints in multi-person occlusion scenarios and \cite{qiu2020peeking} introduced a graph based approach for the task. \cite{wang2022ocr} proposed an occlusion-aware approach that predicts 3D human poses from available ground-truth 2D keypoints, reducing reliance on expensive labeled 3D data but still requiring labeled 2D poses. Despite these advancements, the effectiveness of these methods depends on supervised learning, which in turn requires costly annotations. \\

\noindent \textbf{UDA for Human Pose Estimation:} To address the challenge of limited generalizability in current human pose estimation algorithms across unseen domains, RegDA~\cite{jiang2021regressive} introduced an adversarial learning-based unsupervised domain adaptation (UDA) algorithm. UniFrame~\cite{kim2022unified} employed a Mean-Teacher approach for pseudo label refinement. \cite{cao2019cross} proposed a UDA algorithm for anaimal pose estimation. Additionally, source-free UDA methods proposed in~\cite{peng2023source, raychaudhuri2023prior, ding2023maps} have significantly enhanced the accuracy of current methods on out-of-domain images. However, these algorithms still struggle  when the human figure is partially obscured by occlusions. 

In contrast to these existing approaches, we focus on the problem of predicting 2D human poses under occlusions within unsupervised domain adaptive settings. We argue that current UDA algorithms for this task~\cite{jiang2021regressive, kim2022unified} perform sub-optimally when the human subject in the unlabeled target domain data is occluded.

%% file: sections/3_method.tex
\begin{figure*}[!htb]
    \centering
    \includegraphics[width = 1\textwidth]{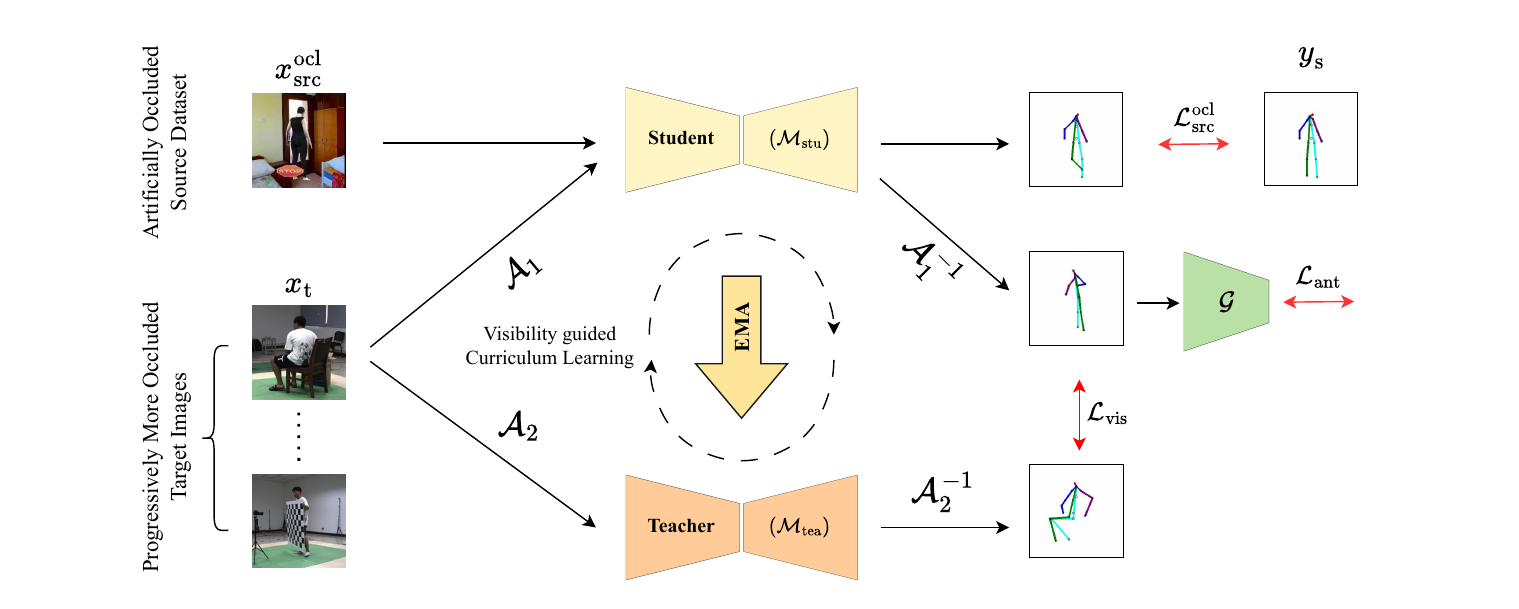}
    \caption{\textbf{Overview of proposed methodology:} \ours~is built upon the mean-teacher framework, where the weights of the teacher model are updated as an exponential moving average (EMA) of the student model's weights. 
    \ours~uses occlusion augmentations on the source domain, enabling the student and teacher models to provide better pseudo labels on unlabeled target images by learning consistency between occluded and unoccluded source images. \ours~also utilizes a human pose prior that captures plausible human anatomy as a zero-level set, thus penalizing anatomically implausible predictions. Finally, \ours~leverages a visibility-based curriculum learning strategy wherein the model focuses on samples with fewer occlusions and gradually shifts to samples with higher degrees of occlusion.} 
    \label{fig:methodology}
\end{figure*}

\section{Methodology}

In this section, we elaborate upon our proposed algorithm (\ours) for unsupervised domain adaptation for occlusion resilient human pose estimation. Our goal is to develop a human pose estimation model $\mathcal{F}$ that takes an image $x \in \mathcal{R}^{H \times W \times 3}$ with spatial dimensions $H$ and $W$ as input and predicts the 2D pose keypoints $y = \mathcal{F}(x)$, where $y \in \mathcal{R}^{K \times 2}$ for $K$ keypoints. We assume access to a source dataset $\mathcal{S} = \{x_{i}, y_{i} \}_{i=1}^{N_{1}}$, where $x_{i}$ is an RGB image and $y_{i}$ is the corresponding set of pose keypoints. We are also provided with an unlabeled target dataset $\mathcal{T}$, containing images $\{ x_{i} \}_{i=1}^{N_{2}}$ without pose keypoint annotations. For clarity, we refer to an image from the labeled source dataset as $x_{s}$, with $y_{s}$ being the corresponding set of pose keypoints, and an image from the unlabeled target set as $x_{t}$. We aim to adapt a model trained on the source dataset, denoted as $\mathcal{F_{\mathcal{S}}}$, to enhance its performance on the target dataset $\mathcal{T}$. The model under adaptation/adapted model is denoted as $\mathcal{F_{\mathcal{T}}}$. 

We propose an intuitive self-learning algorithm for optimal human pose estimation in unsupervised domain adaptive settings, with a particular focus on situations where the human subject in the unlabeled target data is partially occluded by inanimate objects. Existing domain adaptive algorithms for human pose estimation~\cite{jiang2021regressive, kim2022unified} often suffer from error accumulation due to learning from incorrect pseudo labels and a lack of contextual understanding of the human body. This results in sub-optimal performance when the human subject is partially occluded in the unlabeled target data. 

To address these challenges, we introduce three distinct methods: \begin{itemize}

    \item We employ an exponentially weight-averaged teacher model~\cite{tarvainen2017mean} to generate reliable pseudo labels for self-training. This approach ensures that the teacher model does not overfit to incorrect pseudo labels. Additionally, we artificially occlude the labelled source images, this acts as an augmentation step thereby allowing the student model to learn estimating human poses under occlusions.

    \item A human pose prior~\cite{tiwari22posendf}, designed as a zero-level set for plausible human poses, is utilized to regularize the adaptation process by penalizing the model for anatomically implausible predictions.

    \item Recognizing that not all unlabeled target images are equally occluded, we enable the network under adaptation to initially focus on samples with less relatively occlusions. As the adaptation progresses, the network gradually addresses more heavily occluded samples. This approach prevents the network from overfitting to incorrect pseudo labels from the more heavily occluded samples. 
\end{itemize}

\subsection{Pre-Training on Source Data}

In line with previous approaches for human pose estimation, we adopt~\cite{toshev2014deeppose}, thus treating the problem as a heatmap regression task. Additionally, following existing domain adaptive pose estimation works~\cite{jiang2021regressive, kim2022unified, raychaudhuri2023prior, peng2023source}, the weights of $\mathcal{F_{T}}$ are initialized using the weights from a model that has been trained on labeled data from the source domain through supervised learning. This model, known as the source model ($\mathcal{F_{S}}$), is trained by minimizing the loss function defined in Eqn.~\ref{loss:sup}. 

\begin{equation}
\mathcal{L}_{\text{src}} = \frac{1}{N_1} \sum_{x_s} \| \mathcal{F_{S}}(x_s) - y_s \|_{2}^{2}\;.
\label{loss:sup}
\end{equation}

\subsection{Self-Training and Occlusion Augmentations}

We leverage the Mean-Teacher framework~\cite{tarvainen2017mean} for our problem settings. In the Mean-Teacher framework, two identical models are utilized: $\mathcal{M}_{stu}$ (student) and $\mathcal{M}_{tea}$ (teacher), with parameters denoted as $\theta_{stu}$ and $\theta_{tea}$, respectively. Both models are initialized with the weights from a source pre-trained model ($\mathcal{F_{S}}$). During each adaptation step $t$, the student model’s parameters ($\theta_{stu}$) are updated through backpropagating the gradient computed at that step, while the teacher model’s parameters ($\theta_{tea}$) are adjusted as an exponential moving average of the student model’s weights ($\theta_{stu}$) and as a function of smoothing parameter $\alpha$ as outlined in Eqn.~\ref{eqn:ema}. This approach aims to improve overall model performance by reducing overfitting to potentially incorrect pseudo labels, particularly during the initial phase of adaptation.
\begin{equation}
    \theta_{tea}^{t} \longleftarrow \alpha\theta_{tea}^{t-1} + (1-\alpha)\theta_{stu}^{t}
    \label{eqn:ema}
\end{equation} 
We enforce a prediction space consistency for unlabelled target target images across two different augmentations. For any sampled target image $x_{t}$, we apply augmentations $\mathcal{A}_{1}(.)$ and $\mathcal{A}_{2}(.)$ on the same, resulting in augmented images $\mathcal{A}_{1}(x_{t})$ and $\mathcal{A}_{2}(x_{t})$. Corresponding heatmap predictions are obtained as $\mathcal{H}_{1}$ = $\mathcal{M}_{stu}({\mathcal{A}_{1}(x_{t})})$ and $\mathcal{H}_{2}$ = $\mathcal{M}_{tea}({\mathcal{A}_{2}(x_{t})})$. The pseudo labels are obtained by extracting the pixel coordinates with the highest activations in the heatmap predictions of the teacher model ($\mathcal{H}_{2}$). In order to avoid overfitting to potentially incorrect pseudo labels predicted by the teacher model, \ours~only learns from predictions that exceed a specified confidence threshold ($\tau$)~\cite{kim2022unified}. These confident pseudo labels are then converted into normalized heatmaps, which are used to update $\mathcal{M}_{stu}$ by minimizing the loss function defined in Eqn.~\ref{eqn:pred}.

\begin{equation} 
\label{eqn:pred}
\begin{aligned}
    \mathcal{L}_{\text{pred}} &= \frac{1}{|\mathcal{B}|} \sum_{x_{t} \in \mathcal{B}} \sum_{k \in K} \mathbf{1}(\mathcal{H}_{2}^{k} \geq \tau) \\
    &\quad \left\| \mathcal{A}_1^{-1}(\mathcal{H}_{1}^{k}) - \mathcal{A}_2^{-1}(\mathcal{H}_{2}^{k}) \right\|_2^2.
\end{aligned}
\end{equation}
Here, $\mathcal{A}_{1}^{-1}(.)$ and $\mathcal{A}_{2}^{-1}(.)$ are defined as inverse augmentations of $\mathcal{A}_{1}(.)$ and $\mathcal{A}_{2}(.)$ respectively and $\mathcal{B}$ denotes a given batch of images.

During this self training process, we also artificially occluded the labelled source domain images ($x_{s}$) with objects from the COCO dataset~\cite{lin2014microsoft}, referring to these images as ${x_{src}^{ocl}}$. This allows the student model to generalize better to occlusions by leveraging the labeled source data and minimizing the loss function defined by Eqn.~\ref{loss:src-ocl}. 

\begin{equation}
\mathcal{L}_{src}^{ocl} = \frac{1}{|\mathcal{B}|} \sum_{x_{src}^{ocl} \in \mathcal{B}} \| \mathcal{M}_{stu}(x_{src}^{ocl}) - y_s \|_{2}^{2}\;.
\label{loss:src-ocl}
\end{equation}

We observe that leveraging the loss functions described by Eqn.~\ref{eqn:pred} and Eqn.~\ref{loss:src-ocl} leads to improved performance over UniFrame~\cite{kim2022unified}. However, the adapted model may still produce anatomically implausible poses due to a lack of human anatomical constraint, thus leading to suboptimal overall performance.

\subsection{Regularization with Human Pose Prior}

To incorporate an anatomical constraint in the adaptation process, we propose leveraging a pretrained human pose prior. 
We design this parametric prior as a zero-level set for plausible human poses. Specifically, we represent 2D poses as a set of connections (bones) between two joints on the human body. Mathematically, any given skeleton $y$ may be represented by a set of vectors that connect two different joints on the human body, 
\begin{equation}
    \theta_{y} = \{b_{1}, b_{2}, \ldots, b_{M}\}; \quad b_{m} \in \mathcal{R}^{2} \quad \forall m \in [M]
\end{equation} 

We propose to train the parametric human pose prior ($\mathcal{G}: \theta_{y} \rightarrow \mathcal{R}^{+}$) such that 
\begin{equation}
    \mathcal{G}(\theta_{y})
    \begin{dcases}
         = 0, & \theta_{y} \text{ is plausible;} \\
         > 0, & \text{otherwise.}
    \end{dcases}
\end{equation}

We define $\mathcal{G} = \mathcal{G}_{dec} \circ \mathcal{G}_{enc}$, where $\mathcal{G}_{enc}$ projects individual vectors ($b_{m}$) to a latent space based on the anatomical structure of the human body~\cite{mihajlovic2021leap} and $\mathcal{G}_{dec}$ projects this feature in the latent space to a non-negative scalar representing the distance of the queried pose from the learned manifold.

Training $\mathcal{G}$ is accomplished by leveraging the labeled source dataset ($\mathcal{S}$), which contains $N_{1}$ accurate human poses. However, the source dataset ($\mathcal{S}$) does not include examples of implausible human poses, which are also important for learning the pose prior $\mathcal{G}$. To address this, we propose generating examples of implausible poses by artificially occluding the images in $\mathcal{S}$ (\cite{sarandi2020robust}) and estimating the pose using the $\mathcal{F}_{S}$ model that pretrained on clean, unoccluded images of $\mathcal{S}$.

For any image from the labeled source dataset (denoted as $x_{s}$), let $y_{s}$ be the corresponding ground-truth pose label. Let $\hat{x_{s}}$ denote the image generated by artificially occluding $x_{s}$. The corresponding noisy pose estimates are obtained as $\hat{y_{s}} = \mathcal{F}_{S}(\hat{x_{s}})$. Additionally, following~\cite{tiwari22posendf}, we also generate a set of implausible poses by adding noise sampled from the Von-Mises distribution~\cite{gatto2007generalized} to the individual keypoints.

We denote the distance between the plausible ground-truth pose ($y_{s}$) and the correspondingly generated implausible pose ($\hat{y_{s}}$) as \textbf{d}. We set \textbf{d} = 0 if $\theta_{y}$ is sampled from the distribution of poses encompassed by the source dataset $\mathcal{S}$, and \textbf{d} $>$ 0 otherwise.

To train $\mathcal{G}$, we utilize supervised learning by minimizing the $L_{2}$ loss between the predicted score for a given pose $\theta_{y}$ and the corresponding ground-truth score. Mathematically, this can be expressed as: 
\begin{equation}
    \argmin_{\mathcal{G}} \quad ||\mathcal{G}(\theta_{y}) - \textbf{d}||_{2}^{2}
\end{equation}

During adaptation, \ours~leverages this trained human pose prior as a regularizer for the anatomical plausibility of the predicted poses on the unlabelled target domain images ($x_{t}$) 
Mathematically, we minimize: 
\begin{equation}
\mathcal{L}_{ant} = \frac{1}{|\mathcal{B}|} \sum_{x_t \in \mathcal{B}} \mathcal{G}(\mathcal{M}_{stu}(x_t)).
\label{loss:ant}
\end{equation}


\subsection{Visibility guided Curriculum Learning}

The performance of our pose estimation model under adaptation ($\mathcal{F}_{T}$) improves on occluded target domain images by adapting across the prediction space (Eqn.~\ref{eqn:pred} and Eqn.~\ref{loss:src-ocl}) and incorporating a human pose prior to regularize the predicted poses (Eqn.~\ref{loss:ant}). However, since real-world images often vary in their levels of occlusion, we hypothesize that adapting $\mathcal{F}_{T}$ using unlabelled target domain images with fewer occlusions initially, and gradually progressing to images with more occlusions, will enhance $\mathcal{F}_{T}$'s performance by enabling it to learn from more robust and accurate pseudo labels. To implement this, we propose a novel curriculum learning strategy~\cite{karim2023c} that incrementally adapts the student model ($\mathcal{M}_{stu}$) from images with less occlusion to those with relatively more occlusions.

To condition the adaptation process based on human visibility in occluded scenes, we propose using silhouette segmentation maps from the unlabelled target domain images. For given images in $\mathcal{T}$ we can extract binary silhouettes of the visible portions of the humans using pretrained models such as~\cite{li2020self, chen2017rethinking}. For all unlabeled target images from $\mathcal{T}$, we extract binary silhouettes $\{ s_{i} \}_{i=1}^{|\mathcal{T}|}$ using an off-the-shelf pretrained segmentation model ($\mathbf{S}$). Mathematically, $ s_{i} = \mathbf{S}(x_{i})$ $\forall$ $x_{i} \in \mathcal{T}$. Leveraging these extracted silhouettes $\{ s_{i} \}_{i=1}^{|\mathcal{T}|}$, we define visibility score ($v_{i}$) for a given image in batch ($\mathcal{B}$) sampled from $\mathcal{T}$ as: \begin{equation}
    v_{i} = sum(s_{i})/max(sum(s_{i})) \quad \forall s_{i} \in \mathcal{B}
\end{equation} 
Here, $sum(.)$ operator is defined a the total count of foreground pixels in the visible silhouette. Clearly $v_{i} \ \in [0,1]$, with a higher score reflecting that the sample is relatively less occluded. We use these visibility scores to guide the adaptation algorithm by weighing the contributions of individual images in a given batch ($\mathcal{B}$) sampled from $\mathcal{T}$. Thus, more visible images (hence, higher $v_{i}$) will have a larger contribution in the $\mathcal{L}_{\text{pred}}^{\text{vis}}$ (Eqn.~\ref{eqn:pred-vis}), thereby forcing to the model to learn from those samples especially during the initial stages of adaptation. To this end, we propose
\begin{equation} 
\label{eqn:pred-vis}
\begin{aligned}
    \mathcal{L}_{\text{pred}}^{\text{vis}} &= \frac{1}{|\mathcal{B}| \times \sum v_{i}} \sum_{x_{t} \in \mathcal{B}} \sum_{k \in K} \mathbf{1}(\mathcal{H}_{2}^{k} \geq \tau) \\
    &\quad \cdot v_{i} \left\| \mathcal{A}_1^{-1}(\mathcal{H}_{1}^{k}) - \mathcal{A}_2^{-1}(\mathcal{H}_{2}^{k}) \right\|_2^2.
\end{aligned}
\end{equation}

Observe that solely leveraging the loss described by Eqn.~\ref{eqn:pred-vis} will largely ignore heavily occluded images from $\mathcal{T}$. This is not desirable as this might lead to suboptimal performance on heavily occluded images in $\mathcal{T}$. Thus, we resort to using a linear combination of the loss functions defined by Eqn.~\ref{eqn:pred-vis} and Eqn.~\ref{eqn:pred}. We define \begin{equation}
    \mathcal{L}_{vis} = \gamma \mathcal{L}_{pred}^{vis} + (1 - \gamma)  \mathcal{L}_{pred}
    \label{loss:vis}
\end{equation}
with $\gamma = \exp{(\frac{-current \ epoch}{total\ epochs})}$. Thus, during the initial stages of the adaptation $\gamma \rightarrow 1$, ensures that $ \mathcal{L}_{vis}$ focuses more on the samples with relatively less occlusions while gradually as the adaptation progresses, the model focuses equally on all samples in a given batch ($\mathcal{B}$). 
\subsection{Final Objective}
Leveraging the aforementioned loss functions, the student model ($\mathcal{M}_{stu}$) is trained using the weighted objective function, which is mathematically defined as:
\begin{equation}
\argmin_{\mathcal{M}_{stu}} \quad \mathcal{L}_{src}^{ocl} + \lambda_{a} \mathcal{L}_{ant} + \lambda_{v} \mathcal{L}_{vis}
\label{loss:overall}
\end{equation}

%% file: sections/4_experiments.tex
\section{Experiments and Results}

In this section, we showcase \ours's exceptional capability in estimating human poses under occlusions across various unlabeled target datasets. We offer both qualitative and quantitative analyses across several benchmarks, and present an ablation study that emphasizes the significance of each component within our framework. \\

\subsection{Datasets:} 

We utilize the \textbf{SURREAL} dataset~\cite{varol2017learning} for all our experiments due to its extensive collection of over 6 million synthetic images depicting artificial humans, generated from 3D sequences of human motion in an indoor environment. Our primary target dataset is \textbf{3DOH50K}~\cite{zhangoohcvpr20}, which comprises more than 50,000 images of humans in various occluded scenarios within an indoor setting. For our evaluations, we used 12,400 images, with the remaining images allocated for adaptation. Given the scarcity of datasets featuring occluded humans, we create synthetically occluded datasets using \textbf{Humans3.6m (H36M)}~\cite{h36m_pami} and \textbf{Leeds Sports Pose (LSP)}~\cite{johnson2010clustered}, following the approach in~\cite{zhou2021human, dutta2024poise}, by adding occlusion objects from the Pascal VOC dataset~\cite{everingham2010pascal}. This allows us to have the ground-truth unoccluded poses for quantitative evaluation. We refer to these artificially occluded datasets as \textbf{Ocl-H36M} and \textbf{Ocl-LSP}, respectively. The adaptation and evaluation splits for both \textbf{Ocl-H36M} and \textbf{Ocl-LSP} are identical to those used for \textbf{H36M} and \textbf{LSP} in RegDA~\cite{jiang2021regressive} and UniFrame~\cite{kim2022unified}. \\

\subsection{Implementation Details:} For our pose estimation model, we utilize the ResNet101~\cite{he2016deep} pretrained on the ImageNet dataset in combination with the Simple Baseline decoder~\cite{xiao2018simple} as the feature extractor $\mathcal{P}$, following~\cite{jiang2021regressive, kim2022unified}. Augmentations such as rotation, translation and shear as specified in \cite{kim2022unified} were used in this work. The hyperparameter $\tau$ in Eqn.~\ref{eqn:pred} and Eqn.~\ref{eqn:pred-vis}, with set to 0.5, based on the UniFrame~\cite{kim2022unified}. For the loss function in Eqn.~\ref{loss:overall}, we used $\lambda_{a} = 1 \times 10^{-5}$ and $\lambda_{v} = 1$ across all experiments. Consistent with~\cite{tiwari22posendf}, the human pose prior model $\mathcal{G}$ is implemented as a 7-layer MLP network, comprising two encoding layers and five decoding layers. All models were trained with a batch-size of 32 and 500 iterations for epoch for a total of 70 epochs with the Adam optimizer~\cite{kingma2014adam} with an initial learning rate of 1e-04, decaying by a factor of 0.1 after the 45$^{th}$ and 60$^{th}$ epochs. PyTorch was used as the coding framework and all experiments were conducted on a single 24 GB NVIDIA RTX 3090. \\

\noindent \textbf{Baselines:} We primarily evaluate \ours's performance against two state-of-the-art algorithms for human pose estimation: RegDA~\cite{jiang2021regressive} and UniFrame~\cite{kim2022unified}. Additionally, we report the performance of the {\it Source only} model, which serves as a lower bound. This refers to the model's performance on the unlabeled target data when trained exclusively on labeled source data. \\

\noindent \textbf{Metrics:} In line with~\cite{jiang2021regressive, kim2022unified}, we report PCK@0.05 for twelve joints: left and right shoulders (Sld.), elbows (Elb.), wrists, hips, knees, and ankles, as well as their average. PCK@0.05 measures the percentage of correct keypoint predictions within 5\% of the image size. \\

\begin{table}[!htb]
    \centering
    \resizebox{\columnwidth}{!}{%
    \begin{tabular}{lcccccccc}
        \toprule
       Algorithm & Sld. & Elb. & Wrist & Hip & Knee & Ankle & Avg. \\
        \midrule
        \textit{Source only} & 74.7 & 62.5 & 50.4 & 41.2 & 57.9 & 56.0 & 57.1 \\
        RegDA~\cite{jiang2021regressive} & \textbf{85.5} & 73.3 & 56.5 & 50.5 & 67.9 & 71.0 & 67.5 \\
        UniFrame~\cite{kim2022unified} & 55.3 & 75.5 & 68.2 & 68.0 & 78.9 & 69.4 & 69.2 \\
        \ours~(\textbf{ours}) & 78.7 & \textbf{81.4} & \textbf{72.2} & \textbf{70.7} & \textbf{82.2} & \textbf{76.4} & \textbf{76.9} \\
        \bottomrule
    \end{tabular}
    }
    \caption{PCK@0.05 across different joints for \textbf{SURREAL} $\rightarrow$ \textbf{3DOH50K}. Best results in \textbf{bold}}
    \label{tab:S2oh50k}
\end{table}

\begin{table}[!htb]
    \centering
    \resizebox{\columnwidth}{!}{%
    \begin{tabular}{lcccccccc}
        \toprule
       Algorithm & Sld. & Elb. & Wrist & Hip & Knee & Ankle & Avg. \\
        \midrule
        \textit{Source only} & 49.1 & 49.3 & 37.0 & 26.1 & 41.3 & 41.1 & 40.7 \\
        RegDA~\cite{jiang2021regressive} & 68.7 & 73.1 & 53.7 & 52.7 & 66.7 & 69.1 & 64.0\\
        UniFrame~\cite{kim2022unified} & 69.7 & 77.8 & 61.4 & 57.7 & 70.9 & 64.5 & 67.0 \\
        \ours~(\textbf{ours}) & \textbf{73.2} & \textbf{85.0} & \textbf{68.4} & \textbf{67.8} & \textbf{79.2} & \textbf{75.0} & \textbf{74.8} \\
        \bottomrule
    \end{tabular}
    }
    \caption{PCK@0.05 across different joints for \textbf{SURREAL} $\rightarrow$ \textbf{Ocl-H36M}. Best results in \textbf{bold}.}
    \label{tab:S2oclh36m}
\end{table}

\begin{table}[!htb]
    \centering
    \resizebox{\columnwidth}{!}{%
    \begin{tabular}{lcccccccc}
        \toprule
       Algorithm & Sld. & Elb. & Wrist & Hip & Knee & Ankle & Avg. \\
        \midrule
        \textit{Source only} & 44.7 & 53.5 & 49.5 & 43.5 & 40.7 & 45.2 & 46.2 \\
        RegDA~\cite{jiang2021regressive} & 55.0 & 65.0 & 55.8 & 66.6 & 59.8 & 61.6 & 60.6 \\
        UniFrame~\cite{kim2022unified} & 58.0 & 72.1 & 65.3 & 68.3 & 55.2 & 60.7 & 63.3 \\
        \ours~(\textbf{ours}) & \textbf{59.5} & \textbf{75.7} & \textbf{67.8} & \textbf{74.7} & \textbf{65.2} & \textbf{70.2} & \textbf{68.9} \\
        \bottomrule
    \end{tabular}
    }
    \caption{PCK@0.05 across different joints for \textbf{SURREAL} $\rightarrow$ \textbf{Ocl-LSP}. Best results in \textbf{bold}.}
    \label{tab:S2oclLSP}
\end{table}

\subsection{Quantitative Results:} In Tables~\ref{tab:S2oh50k},~\ref{tab:S2oclh36m}, and~\ref{tab:S2oclLSP}, we present a comparative quantitative analysis of our proposed algorithm, \ours, against the state-of-the-art baseline algorithms, RegDA~\cite{jiang2021regressive} and UniFrame~\cite{kim2022unified}, across three benchmark settings: SURREAL $\rightarrow$ 3DOH50K, SURREAL $\rightarrow$ Ocl-H36M, and SURREAL $\rightarrow$ Ocl-LSP respectively. These target datasets are characterized by significant occlusions, and the fact that \ours~consistently outperforms existing algorithms by $\approx$ 7\% across all three benchmarks demonstrates its effectiveness in handling occlusions within unsupervised domain adaptation settings. 

\begin{table}[!htb]
    \centering
    \resizebox{\columnwidth}{!}{%
    \begin{tabular}{lccccccc}
        \toprule
        Algorithm & Sld. & Elb. & Wrist & Hip & Knee & Ankle & Avg. \\
        \midrule
        \textit{Source only} & 51.5 & 65.0 & 62.9 & 68.0 & 68.7 & 67.4 & 63.9 \\
        RegDA~\cite{jiang2021regressive} & 62.7 & 76.7 & 71.1 & 81.0 & 80.3 & 75.3 & 74.6 \\
        UniFrame~\cite{kim2022unified} & \textbf{69.2} & \textbf{84.9} & \textbf{83.3} & 85.5 & \textbf{84.7} & \textbf{84.3} & \textbf{82.0} \\
        \ours~(\textbf{ours}) & 67.1 & 81.9 & 78.9 & \textbf{85.9} & 83.6 & 82.5 & 80.0 \\
        \bottomrule
    \end{tabular}
    }
    \caption{PCK@0.05 across different joints for \textbf{SURREAL} $\rightarrow$ \textbf{LSP}. Best results in \textbf{bold}.}
    \label{tab:S2LSP}
\end{table}

In Table~\ref{tab:S2LSP}, our analysis of the SURREAL $\rightarrow$ LSP benchmark shows that when the target dataset is free from occlusions, \ours~performs on par with the state-of-the-art UniFrame algorithm~\cite{kim2022unified} and outperforms RegDA~\cite{jiang2021regressive} by $\approx$ 5\%. This indicates that \ours~provides more reliable pose estimates under occlusions while maintaining accuracy comparable to that of existing algorithms~\cite{kim2022unified} in unoccluded scenarios. \\

\begin{figure}[!htb]
    \centering
    \includegraphics[width = 1\linewidth]{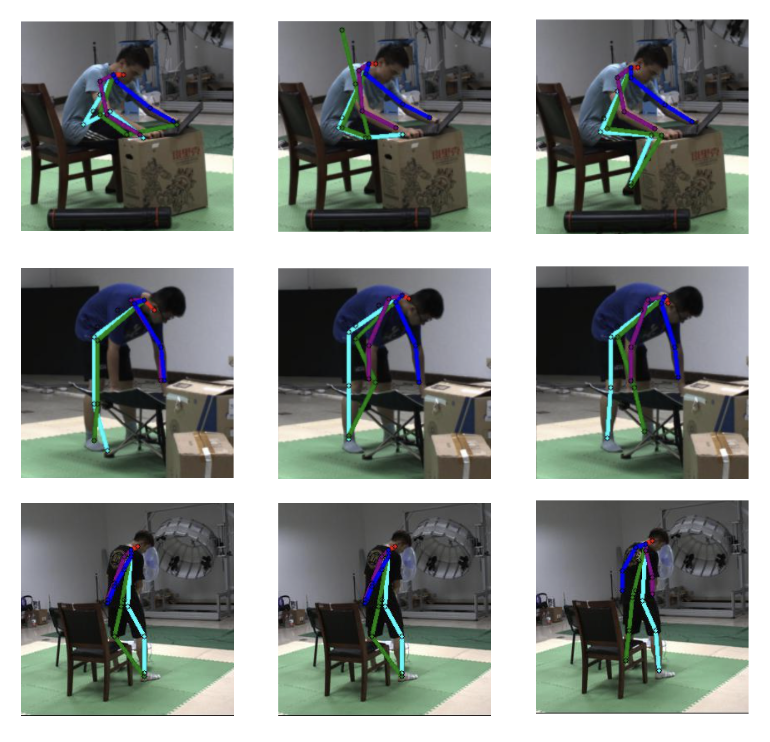}
    \caption{\textbf{Qualitative Results for \textbf{SURREAL}$ \rightarrow$ \textbf{3DOH50K}}. \textit{From left to right:} {\it Source only} predictions, prediction from UniFrame~\cite{kim2022unified}, and predictions our proposed algorithm~\ours.}
    \label{fig:surreal-3doh50k}
\end{figure}

\begin{figure}[!htb]
    \centering
    \includegraphics[scale = 0.6]{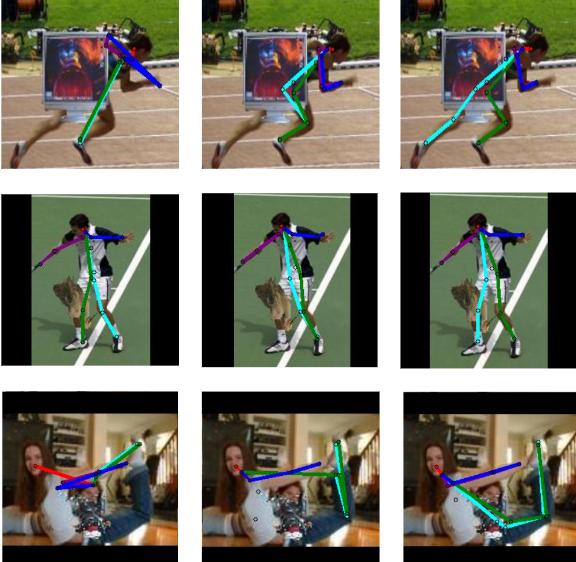}
    \caption{\textbf{Qualitative Results for \textbf{SURREAL}$ \rightarrow$ \textbf{Ocl-LSP}}. \textit{From left to right:} {\it Source only} predictions, prediction from UniFrame~\cite{kim2022unified}, and predictions our proposed algorithm~\ours.}
    \label{fig:surreal-ocllsp}
\end{figure}

\begin{figure}[!htb]
    \centering
    \includegraphics[scale = 0.6]{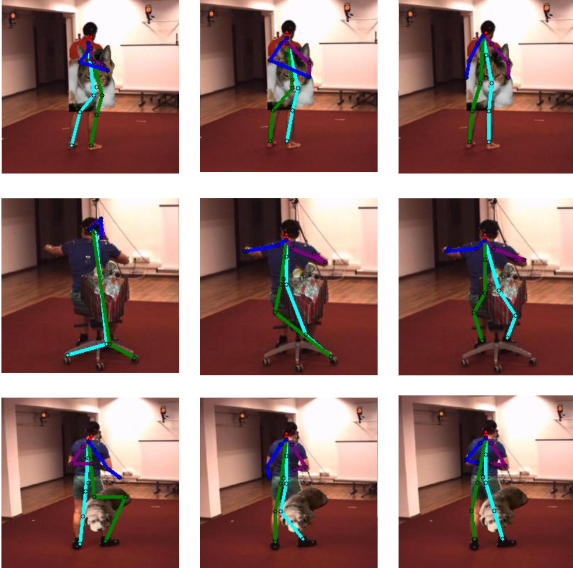}
    \caption{\textbf{Qualitative Results for \textbf{SURREAL}$ \rightarrow$ \textbf{Ocl-H36M}}. \textit{From left to right:} {\it Source only} predictions, prediction from UniFrame~\cite{kim2022unified}, and predictions our proposed algorithm~\ours.}
    \label{fig:surreal-oclh36m}
\end{figure}

\begin{figure}[!htb]
    \centering
    \includegraphics[scale = 0.6]{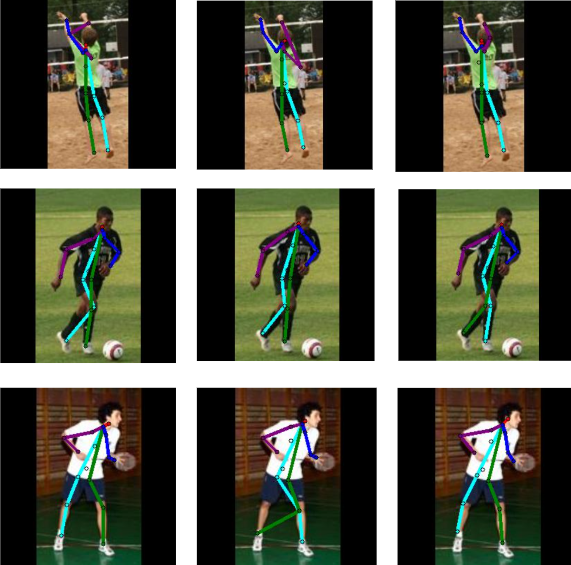}
    \caption{\textbf{Qualitative Results for \textbf{SURREAL}$ \rightarrow$ \textbf{LSP}}. \textit{From left to right:} {\it Source only} predictions, prediction from UniFrame~\cite{kim2022unified}, and predictions our proposed algorithm~\ours.}
    \label{fig:surreal-lsp}
\end{figure}

\subsection{Qualitative Results:} We present qualitative results on the following benchmarks: SURREAL $\rightarrow$ 3DOH50K, SURREAL $\rightarrow$ Ocl-LSP, SURREAL $\rightarrow$ Ocl-H36M and SURREAL $\rightarrow$ LSP, as illustrated in Figures~\ref{fig:surreal-3doh50k}, \ref{fig:surreal-ocllsp}, \ref{fig:surreal-oclh36m} and \ref{fig:surreal-lsp} respectively. It is evident that UniFrame~\cite{kim2022unified} performs poorly in estimating human pose when there are occlusions in unlabeled target images. Specifically, for the SURREAL $\rightarrow$ 3DOH50K benchmark (Figure~\ref{fig:surreal-3doh50k}), UniFrame struggles with natural occlusions, such as those caused by boxes or chairs, resulting in anatomically implausible predictions, such as placing the left ankle where the right ankle should be. Similarly, in the SURREAL $\rightarrow$ Ocl-LSP benchmark (Figure~\ref{fig:surreal-ocllsp}), UniFrame demonstrates comparable inaccuracies, raising concerns about its practical applicability. In contrast, \ours~consistently produces accurate pose estimates, highlighting its effectiveness in handling occlusions in unsupervised domain adaptation scenarios.\\
\begin{table}[!htb]
\centering
\begin{tabular}{ccccccc}
\toprule
$\mathcal{L}_{src}^{ocl}$ & $\mathcal{L}_{pred}$ & $\mathcal{L}_{ant}$ & $\mathcal{L}_{vis}$ & Avg. \\
\hline
 \xmark & \xmark & \xmark & \xmark & 57.1 \\
 \cmark & \cmark & \xmark & \xmark & 72.5 \\
 \cmark & \cmark & \cmark & \xmark & 74.2 \\
 \cmark & - &\cmark & \cmark & \textbf{76.9} \\
\hline
\end{tabular}
\caption{Ablation study for individual components of our proposed algorithm (\ours) for \textbf{SURREAL} $\rightarrow$ \textbf{3DOH50K}. Note that, we do not explicitly use $\mathcal{L}_{pred}$ in our final objective function (Eqn.~\ref{loss:overall}) as the same is already contained in $\mathcal{L}_{vis}$ (Eqn.~\ref{loss:vis}). Best results in \textbf{bold}.}
\label{tab:abl}
\end{table}

\subsection{Ablation Experiments:} 

\noindent \textbf{Contribution of Individual Components:} In Table~\ref{tab:abl}, we present an ablation study to assess the contribution of individual components of our proposed algorithm on the SURREAL $\rightarrow$ 3DOH50K benchmark. We show that simply applying prediction space consistency on unlabeled target data along with occlusion augmentations on the labeled source data (Eqn.~\ref{eqn:pred} and Eqn.~\ref{loss:src-ocl}) allows \ours~to outperform the state-of-the-art algorithm UniFrame by $\approx$ 3\%. Incorporating regularization with the human pose prior (Eqn.~\ref{loss:ant}) improves this performance to about 5\%. Finally, using visibility-guided curriculum learning (Eqn.~\ref{loss:vis}), \ours~outperforms UniFrame by $\approx$ 7\%, achieving state-of-the-art results for human pose estimation with unlabeled occluded target datasets. \\ 

\begin{table}[!htb]
    \centering
    \resizebox{\columnwidth}{!}{%
    \begin{tabular}{lcccccccc}
        \toprule
       Algorithm & Sev-1 & Sev-2 & Sev-3 & Sev-4 & Sev-5 \\
        \midrule
        UniFrame~\cite{kim2022unified} & 70.6 & 69.0 & 64.3 & 64.0 & 56.9 \\
        \ours~(\textbf{ours}) & \textbf{73.4} & \textbf{71.5} & \textbf{68.9} & \textbf{68.7} & \textbf{64.9} \\
        \bottomrule
    \end{tabular}
    }
    \caption{PCK@0.05 across different severities for \textbf{SURREAL} $\rightarrow$ \textbf{Ocl-LSP}. Best results in \textbf{bold}.}
    \label{tab:sev_comp}
\end{table}

\noindent \textbf{Sensitivity to Occlusion Sizes:} We compare OR-POSE and UniFrame~\cite{kim2022unified} at five increasing severity levels, where the size of occlusions increases with severity as shown in \ref{tab:sev_comp}. At severity level 1, occlusions are approximately 48x48 pixels in 256x256 images, while at severity level 5, they increase to around 96x96 pixels. Our results indicate that as occlusion size increases, the performance of OR-POSE declines, but the degradation in UniFrame’s performance is significantly more pronounced. This demonstrates that OR-POSE remains more robust than UniFrame even under larger occlusions. Consequently, it can be concluded that OR-POSE is substantially more reliable than state-of-the-art algorithms when dealing with occlusions of varying types, shapes, and sizes.

%% file: sections/5_conclusion.tex
\section{Conclusion}

We introduce \ours, a novel self-learning algorithm for occlusion resilient human pose estimation under unsupervised domain adaptive settings. Unlike existing UDA algorithms that generate suboptimal and anatomically incorrect poses under occlusions in unlabeled target datasets, \ours~consistently achieves optimal performance. \ours~utilizes the mean-teacher framework for iterative pseudo-label refinement, combined with a human pose prior that regularizes the algorithm to maintain anatomical plausibility, thereby enhancing performance under occlusions. Further, \ours~implements a visibility-guided curriculum learning approach, aiding the model to initially learn from less occluded images and gradually adapt to more occluded ones, hence reducing error accumulation from incorrect pseudo labels generated by highly occluded samples. We show that \ours~outperforms current methods on challenging benchmarks for occluded human pose estimation while maintaining strong performance on non-occluded benchmarks, showing its versatility in real-world scenarios.